# Probability Estimation and Scheduling Optimization for Battery Swap Stations via LRU-Enhanced Genetic Algorithm and Dual-Factor Decision System

Anzhen Li[1†], Shufan Qing[1†], Xiaochang Li[1], Rui Mao[1] and Mingchen Feng[1*]

**Abstract.** To address the challenges of limited Battery Swap Stations datasets, high operational costs, and fluctuating user charging demand, this research proposes a probability estimation model based on charging pile data and constructs nine scenario-specific battery swap demand datasets. In addition, this study combines Least Recently Used strategy with Genetic Algorithm and incorporates a guided search mechanism, which effectively enhances the global optimization capability. Thus, a dual-factor decision-making based charging schedule optimization system is constructed. Experimental results show that the constructed datasets exhibit stable trend characteristics, adhering to 24-hour and 168-hour periodicity patterns, with outlier ratios consistently below 3.26%, confirming data validity. Compared to baseline, the improved algorithm achieves better fitness individuals in 80% of test regions under the same iterations. When benchmarked against immediate swap-and-charge strategy, our algorithm achieves a peak cost reduction of 13.96%. Moreover, peak user satisfaction reaches 98.57%, while the average iteration time remains below 0.6 seconds, demonstrating good computational efficiency. The complete datasets and optimization algorithm are opensourced at https://github.com/qingshufan/GA-EVLRU.

## 1    Introduction

Electric vehicles (EVs) are an effective solution for achieving the carbon peak and carbon neutrality [1,2]. However, the contradiction between user demands brought about by the popularization of EVs and the existing charging mode has exacerbated problems of "difficulty in charging" and "slow charging". Among them, the battery swapping mode, as a new type of energy replenishment method, can complete battery swap at Battery Swap Station (BSS) within 3-5 minutes. The continuous maturity of the battery swapping mode not only enhances the market-oriented operation potential of BSS but also effectively alleviates the pressure of peak regulation and frequency modulation. At the same time, compared with the charging mode, the battery swapping mode has overwhelming advantages in terms of energy replenishment efficiency, battery wear, space occupation, safety, and waiting time.

Operating framework diagram of the BSS is shown in **Fig. 1**. Currently, there are already commercial companies such as NIO. The second-generation BSS of NIO reserves 5 battery packs with a capacity of 100 kWh (referred to as Battery A) and 8 battery packs with a capacity of 70/75 kWh (referred to as Battery B). The swapping time of a battery is within 5 minutes, and charging to 93% is within 1 hour [3,4]. However, operating costs of BSS are substantial. Therefore, the optimization of it is of great importance. Moreover, due to the fierce competition among BSS, datasets are scarce and most of them are not publicly available [6], which is not conducive to industry research. Therefore, in the initial stage of research, how to reasonably use a large number of charging pile datasets to estimate the demand dataset of BSS is a key issue. Furthermore, a key problem is that the number of fully charged batteries usually reserved at the stations is limited, while the demand quantities of different battery types change dynamically. During peak periods, fully charged batteries of various types will be quickly consumed due to large demand for EV users. In order to continuously meet the needs of users and ensure them a high-quality experience, it is particularly important to study the dynamic scheduling of heterogeneous batteries and the optimization of charging plans in BSS.

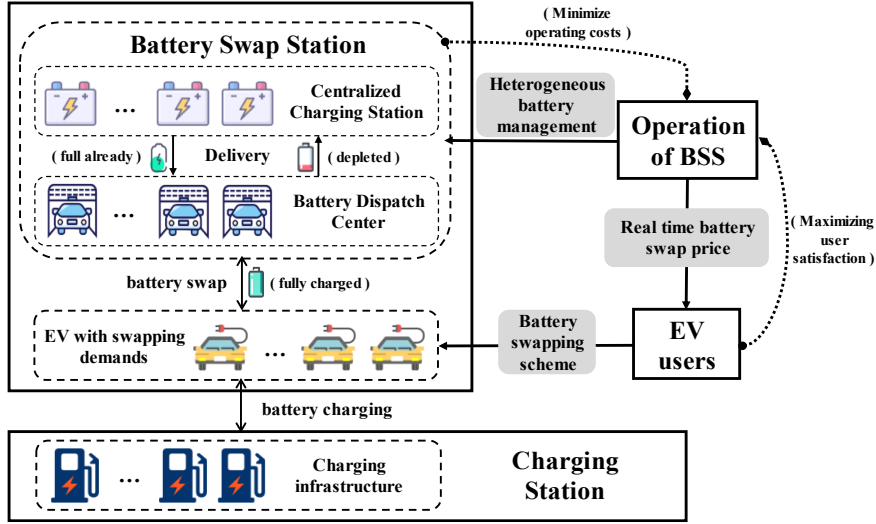

**Fig. 1.** Operating framework diagram of the BSS. When choosing battery swapping, the EV will complete the battery swapping at the Battery Dispatch Center and used batteries will be transported to a Centralized Charging Station for charging. The fully charged batteries will be sent back for subsequent use by EVs, thus realizing the cyclic utilization of battery energy. BSS achieve diversified operation benefits through centralized battery management.

In this study, we construct demand datasets for BSS with heterogeneous batteries and design a composite decision support system for multi-objective collaborative optimization. The main contributions are as follows:

- In view of the industry challenge of scarce BSS datasets, we propose a probability estimation model based on charging pile data and construct nine battery swap demand datasets, providing a foundation for research in this field.
- We innovatively integrate Least Recently Used (LRU) into Genetic Algorithm (GA), significantly enhancing global search ability through guided search mechanism.

• We develop a dual-factor decision-based charging schedule optimization system for a battery swapping station that achieves charging plan formulation with both lower costs and higher user satisfaction.

The rest of the paper is organized as follows: Section 2 reviews related works. Section 3 elaborates and illustrates the proposed algorithm. Section 4 presents the experimental results on the nine datasets. Section 5 concludes the paper.

## 2    Related Work

**Battery Swapping Demand Datasets.** Qu et.al [7] utilizes graphs and temporal attention mechanisms, as well as physics-informed meta, to predict and has publicly released the ST-EVCDP charging pile dataset. Kuang et.al [8] has studied the impact of electricity prices on the charging behavior of EVs. Qu et.al [9] has fine-tuned a Large Language Model using a meta-learning framework for EV charging prediction. Li et.al [10] has made predictions based on the Transformer and publicly released the UrbanEV charging pile dataset. However, most studies focus on the construction of charging station datasets, the data scarcity faced by BSS [4] still have not been well resolved.

**Enhanced GA.** Chen et.al [11] accelerates the operation of GA by ignoring unpromising item sets. Li et.al [12] combines the Sarsa with the GA to obtain a more optimal scheduling scheme. Du et.al [13] introduces the elite strategy to enhance the search efficiency and global optimization ability of GA. However, most studies mainly focus on optimizing the parameters of GA rather than adjusting the search direction of GA.

**Multi-factor Decision-making Optimization.** Mahoor et.al [14] has studied on the multi-objective collaborative operation of BSS based on uncertainty constraint modeling and battery degradation cost optimization. Wang et.al [15] has studied the day-ahead and real-time collaborative optimization strategy for BSS participation in frequency regulation based on the Information Gap Decision Theory and the Boundary Autonomous Selection method. However, most studies focuses on aspects such as EV charging optimization [5], layout optimization [16,17] of BSS, pricing optimization [18], and efficient battery charging plans [19]. There is little research on the optimization of operation costs which are limited to single-objective optimization and often ignore the key factor of user satisfaction.

## 3    Methodology

In this section, we propose a battery swapping demand estimation algorithm based on a probability model. Aiming at the multi-peak characteristics of complex charging scheduling optimization, we combine the inherent advantages of GA in global exploration of the solution space and swarm intelligence optimization, as well as the LRU guiding strategy, and propose the Genetic Algorithm-Electronic Vehicle Least Recently Used (GA-EVLRU) algorithm. The overall framework is shown in **Fig. 2**.

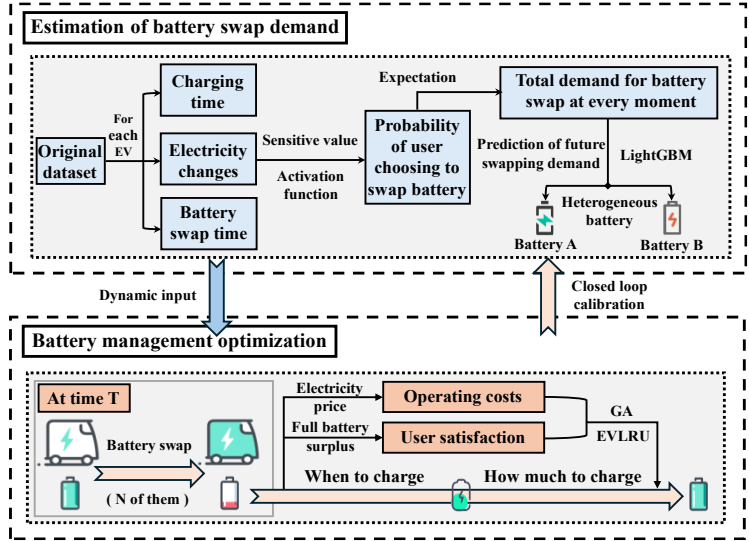

**Fig. 2.** An Overview of the System Architecture Diagram. At the top is the battery swapping demand estimation module, detailed in Section 3.1, and at the bottom is the battery management optimization module, detailed in Section 3.2.

### 3.1 Estimation of Battery Swap Demand Based on a Multi-Factor Probabilistic Model

Users' decisions (whether to charge or swap batteries) primarily depend on the time cost of charging and battery swapping, the total charging energy, and the price. Users who are more sensitive to time costs tend to choose the option with the lower time cost, while users who are more sensitive to electric quantity are more likely to opt for battery swapping when their battery level is low, thereby achieving better economic benefits. By analyzing several publicly available large-scale charging pile datasets [4,5], it is evident that charging time and charging energy can be directly or indirectly calculated. Additionally, NIO [3,4] provides data on the time cost of battery swapping. Based on these considerations, this study establishes a probabilistic model incorporating charging time, battery swapping time, and charging energy to estimate battery swap demand.

$$\begin{cases} z = \boldsymbol{\theta}^T \mathbf{x} = \sum_{i=1}^{3} \theta_i \, x_i \\ P(\mathbf{x}; \boldsymbol{\theta}) = \sigma(\boldsymbol{\theta}^T \mathbf{x}) = \frac{1}{1+e^{-\boldsymbol{\theta}^T \mathbf{x}}} \end{cases} \tag{1}$$

where $\boldsymbol{\theta} = [0.08, 0.08, -0.8]^T$ is the parameter vector, $\mathbf{x} = [t_{sum}, v_{sum}, c_r]^T$ is the feature vector, $t_{sum}$ represents the total charging duration, $v_{sum}$ represents the total charging energy, $c_r$ represents the battery swapping time, $P(\mathbf{x}; \boldsymbol{\theta})$ denotes the probability of battery swap demand given the $\mathbf{x}$ and the $\boldsymbol{\theta}$, and $z$ is an intermediate calculation result.

Assuming that the number of users with battery swap demand depends on the total number of users entering the service area, the expected value of battery swap demand under general conditions can be derived using the Law of Total Expectation as:

$$E(Y) = E\big[E\big(Y|\mathcal{N}(T)\big)\big] = \sum_{n=0}^{\infty} (\sum_{i=1}^{n} P(\mathbf{x}_i; \boldsymbol{\theta})) \, P\big(\mathcal{N}(T) = n\big) \tag{2}$$

where $T$ is the time period, $\mathcal{N}(T)$ represents the total number of users entering the battery swap service area during the $T$ and $n \in \mathbb{N} \cup \{0\}$. $\mathbf{x}_i = [t_{sum}^i, v_{sum}^i, c_r^i]^T$ is the feature vector of the $i$-th user. $Y = \sum_{i=1}^{\mathcal{N}(T)} X_i$ denotes the total number of users with battery swap demand during the $T$, where $X_i$ is a Bernoulli random variable, indicating whether the $i$-th user has battery swap demand.

If $\mathcal{N}(T)$ follows a Poisson distribution with parameter $\lambda$, and the feature vector $\mathbf{x}_i$ are independently and identically distributed (i.i.d.), then the probability of $P(\mathcal{N}(T) = n)$ is as:

$$P(\mathcal{N}(T) = n) = \frac{e^{-\lambda} \lambda^n}{n!}, \quad n = 0,1,2,\cdots \tag{3}$$

Based on the properties of the Poisson distribution and the linearity of expectation, it can be further simplified as:

$$E(Y) = P(\mathbf{x}_i; \boldsymbol{\theta}) \sum_{n=0}^{\infty} n \frac{e^{-\lambda} \lambda^n}{n!} = P(\mathbf{x}_i; \boldsymbol{\theta}) \lambda \tag{4}$$

### 3.2   Optimization of BSS Charging Plan via LRU Heuristic Strategy and GA

This study introduces the heuristic search idea of LRU into the initialization population and mutation operations of GA [11-13]. The specific steps are as follows.

The charging time for Battery A and B is controlled within one hour [3,4]. Therefore, the BSS charging plan designed is based on an hourly unit. The real-time charging plan for Battery A and B over the next 24 hours is encoded as a 48-digit decimal code for individual $I$, where bits 1-24 are for Battery A and bits 25-48 for Battery B.

Let the set of battery packs be $\Omega = \{A, B\}$. For any battery pack $i \in \Omega$, let $m_i$ be its maximum inventory capacity. Then, for the current demand quantity of the specified battery pack, we can calculate the actual supplied quantity $\hat{d}_{i,t}$ as:

$$\hat{d}_{i,t} = \min\big(d_{i,t}, f_{i,t-1}\big) \tag{5}$$

where $t \in T = \{1,2,\cdots,24\}$ is the time step, $d_{i,t}$ represents the demand for the $i$, and $f_{i,t}$ represents the number of fully charged $i$ types at the end of $t$, with $f_{i,0} = m_i$.

The battery quantities are updated at the end of each time step according to the following state transition equations:

$$\begin{aligned} e_{i,t} &= e_{i,t-1} + \hat{d}_{i,t} - c_{i,t} \\ f_{i,t} &= f_{i,t-1} - \hat{d}_{i,t} + c_{i,t} \end{aligned} \tag{6}$$

where $e_{i,t}$ represents the number of empty $i$ types at the end of the $t$, with $e_{i,0} = 0$, and $c_{i,t}$ represents the number of $i$ types charged during the $t$. In a completely random scenario, $c_{i,t}$ is as follows, which implies that the search is blind and directionless:

$$c_{i,t} = U\big(0, e_{i,t}\big) \tag{7}$$

where $U(0, e_{i,t})$ represents a uniform random variable over the interval $[0, e_{i,t}]$.

---

**Algorithm 1:** Individual GA Based on the LRU Strategy

---

**Input:** The demand for battery pack $i$ at time step $t$: $d_{i,t}$, $i \in \Omega$, $j \in \Omega - \{i\}$.

**Output:** An individual representing feasible battery charging solution: $I$.

1:   $f_{i,0} \leftarrow m_i$
2:   $e_{i,0} \leftarrow 0$
3:   **for** t **from** 1 **to** 24 **do**
4:     $\hat{d}_{i,t} \leftarrow min(f_{i,t-1}, d_{i,t})$
5:     $e_{i,t} \leftarrow e_{i,t-1} + \hat{d}_{i,t}$
6:     $f_{i,t} \leftarrow f_{i,t-1} - \hat{d}_{i,t}$
7:     **if** $e_{i,t} > 0$ **then**
8:       $\Delta_{i,t} \leftarrow \max\left(0, \min(d_{i,t} - d_{j,t}, e_{i,t})\right)$
9:       $c_{i,t} \leftarrow U(\Delta_{i,t}, e_{i,t})$
10:    **else**
11:      $c_{i,t} \leftarrow 0$
12:    **end if**
13:    $f_{i,t} \leftarrow f_{i,t} + c_{i,t}$
14:    $e_{i,t} \leftarrow e_{i,t} - c_{i,t}$
15:    $I \leftarrow I \cup \{c_{i,t}\}$

---

For battery pack $i \in \Omega$, let $j \in \Omega$ and $j \neq i$ (representing another battery pack). The demand difference $\Delta_{i,t}$ between different battery packs is as:

$$\Delta_{i,t} = \max\left(0, \min(d_{i,t} - d_{j,t}, e_{i,t})\right) \tag{8}$$

Based on the LRU strategy, we can perform a directed search by updating the charging quantity $c_{i,t}$ according to the demand difference $\Delta_{i,t}$, and the formula is as:

$$c_{i,t} = \begin{cases} U(\Delta_{i,t}, e_{i,t}) & \text{if } e_{i,t} > 0 \text{ and } d_{i,t} > d_{j,t} \\ U(0, e_{i,t}) & \text{if } e_{i,t} > 0 \text{ and } d_{i,t} \leq d_{j,t} \\ 0 & \text{if } e_{i,t} = 0 \end{cases} \tag{9}$$

The individual $I$ generated is a chromosome encoding, which is essentially a vector of length 48. It sequentially stores the charging quantities of battery packs A and B at each time step. Specifically, for $t = 1, 2, \cdots, 24$, when $k = t$, $I_k = c_{A,t}$; when $k = t + 24$, $I_k = c_{B,t}$. The detailed pseudocode for its calculation is shown at **Alg. 1**.

**Design of objective function and fitness function.** To solve the battery management plan, ensuring minimized operational costs while maximizing user satisfaction. Additionally, the number of battery packs taken should not exceed the real-time inventory of battery packs, the number of battery packs charged should not exceed the real-time inventory of empty battery packs, and user satisfaction should not fall below a predefined lower limit. Based on this, the objective function and constraints are as:

$$\min_I f(I) = \frac{\sum_{t \in \{1,2,\cdots,24\}} \left(\sum_{i \in \Omega} c_{i,t}\right) V_t}{\xi_{max}} + \gamma + g(\gamma)$$

$$s.t. \begin{cases} c_{i,t} \leq e_{i,t-1} \\ d_{i,t} \leq f_{i,t-1} \\ \gamma \geq \tau_s \end{cases} \tag{10}$$

where $V_t$ represents the electricity price at time $t$, $\xi_{max}$ represents the maximum cost under the immediate swap-and-charge mode, $\gamma$ represents user satisfaction, $g(\gamma)$ is a penalty term, and $\tau_s$ represents the minimum user satisfaction threshold.

The primary factor affecting user satisfaction $\gamma$ is whether the battery swap demand is met. When the pre-charged battery reserve at the BSS is insufficient, user satisfaction decreases. Based on this principle, the formula for calculating $\gamma$ is designed as:

$$\gamma = \frac{\sum_{i \in \Omega} \sum_{t \in \{1,2,\cdots,24\}} c_{i,t}}{\sum_{i \in \Omega} \sum_{t \in \{1,2,\cdots,24\}} d_{i,t}} \tag{11}$$

Furthermore, to prevent the selected individuals from having the $\gamma$ value below $\tau_s$, a penalty term $g(\gamma)$ is introduced. It effectively eliminates individuals with excessively low $\gamma$ values, thereby obtaining better individuals. The formula is as:

$$g(\gamma) = \begin{cases} 0, & \gamma \geq \tau_s \\ \tau_1, & \gamma < \tau_s \end{cases} \tag{12}$$

where $\tau_1$ is an empirical value, in this paper $\tau_1 = 2$, $\tau_s = 0.9(90\%)$.

**Other Parts Design.** In the selection-copy operation, based on the tournament selection strategy, we randomly select a certain number of individuals (tournament size) from the population and choose the chromosome with the best fitness for replication to preserve individuals representing lower costs and higher $\gamma$ solutions. For the crossover operation, using the midpoint crossover strategy, selected individuals exchange their charging plans for battery packs A and B to generate offspring with potential advantages. Regarding the mutation operation, relying on the random mutation strategy and Monte Carlo simulation, a new individual is directly generated, representing a new solution biased towards the optimal solution based on the LRU strategy.

## 4    Experimental Results

The experiments are divided into three parts. The first part is to verify the effectiveness of the battery swapping demand dataset constructed. The second part is to verify the algorithm proposed. The third part is to formulate the charging plan and conduct benefit evaluation for the actual region. The experimental operating environment is a laptop computer configured as an AMD Ryzen 5 5600H CPU (6-core 3.30 GHz), 16 GB of RAM, and NVIDIA GeForce RTX 3060 Laptop GPU with 6 GB of memory size.

### 4.1    Evaluation of Battery Swap Demand Estimation Algorithm

To evaluate the performance of the battery swapping demand estimation proposed, we have selected three key indicators for quantitative analysis, namely: trend stability ($S_m$) [20], cycle matching degree ($P_m$) [21], and proportion of outliers ($O_m$) [22]. The trend stability $S_m$ (the standard deviation of the slopes of each segment) aims to measure the coherence of the curve's trend within a specific period. If the trend changes frequently, which is contrary to the relatively stable trend of battery swapping demand in the actual scenario, it indicates that the simulation effect is poor. The cycle matching degree $P_m$ is used to evaluate the degree of fit between the simulated and the actual potential periodic characteristics (such as the peak battery swapping during the morning and evening rush hours on working days). The proportion of outliers $O_m$ can characterize the abnormal fluctuations in the simulated curve. The larger the value of $O_m$, the lower the accuracy of the simulation results may be. The formulas are as:

$$S_m = \sqrt{\frac{1}{n}\sum_{i=1}^{n}(s_i - \bar{s})^2} \tag{13}$$

where, $s_i$ represents the slope of the fitted straight line of the curve in each time period, $\bar{s}$ is the mean value of the slopes, and $n$ is the total number of time periods.

**Table 1.** Evaluation results for 9 datasets. We use ↑ for the higher the better and ↓ for the reverse. The best is boldfaced and the second best is underlined. † indicates better than the second best. ✱ indicates that the dataset has charging time. ♦ indicates that the dataset has charging capacity. ✿ indicates that the dataset has electricity prices. ● indicates that the dataset has multiple regions and the average is taken.

| Dataset | $S_m(\times 10^{-3})$ ↓ | $P_m(\times 100\%)$ ↑ | $O_m(\times 100\%)$ ↓ |
|---|---|---|---|
| ST-EVCDP ✱♦✿● | 0.73 | 59.06 | <u>0.94</u> |
| UrbanEV ✱♦✿● | 0.12 | 59.75 | **0.12**† |
| Dundee ♦ | 1.16 | 14.28 | 2.44 |
| Perth ♦ | 0.13 | 80.02 | 2.40 |
| Paris ♦ | 1.51 | 56.92 | 1.04 |
| Boulder ✱♦ | **0.08**† | **99.01**† | 2.88 |
| Palo Alto ✱♦ | 0.18 | <u>99.00</u> | 1.31 |
| SAP ✱♦ | <u>0.09</u> | 80.02 | 2.56 |
| ACN ✱♦ | 0.31 | 80.03 | 2.86 |

$$P_m = \frac{\sum_{i=1}^{n}(a_i b_i)}{\sqrt{\sum_{i=1}^{n} a_i^2}\sqrt{\sum_{i=1}^{n} b_i^2}} \tag{14}$$

where $a_i$ and $b_i$ represent the $i$-th elements in the simulated periodicity list and the expected periodicity list respectively. The expected periods are set to 24 hours and 168 hours, corresponding to one day and one week respectively. (The $P_m$ is obtained by performing a Discrete Fourier Transform (DFT) on the curve to get its frequency domain representation, determining the main periodicity, and then calculating the cosine similarity with the expected periodicity list.)

$$O_m = \frac{1}{n}\sum_{j=1}^{n} R_j$$
$$R_j = \begin{cases} 1, & X_j \notin (\mu - 3\sigma, \mu + 3\sigma) \\ 0, & \text{else} \end{cases} \tag{15}$$

where $\mu$ is the mean value of the sequence $\mathbf{X} = \{X_j\}_{j=1}^{n}$, and $\sigma$ is the standard deviation. The formula is based on the assumption that the estimated sequence data follows a normal distribution.

The datasets used consist of three categories: The ST-EVCDP series [7-9] covers 18,061 public charging piles in Shenzhen, China, spanning from June 19 to July 18, 2022; the UrbanEV [10] series covers 1,362 charging stations and 17,532 charging piles, spanning from September 1, 2022 to February 28, 2023; the EV Load Open Data series comprises datasets from eight regions such as Dundee (UK), Perth & Kinross (UK), and Palo Alto (California, USA). Nine subsets capable of directly or indirectly deriving charging duration and energy consumption are selected, with corresponding evaluation metrics presented in **Table 1**. The $S_m$ values of all nine datasets are very small, which indicates that the battery swap demand curves estimated by our algorithm demonstrate good trend stability. With the exception of the Dundee dataset, the $P_m$ values of the remaining eight datasets all exceed 50%, suggesting that their battery swap demand generally follows 24-hour or 168-hour periodicity. Notably, both Boulder and Palo Alto datasets show $P_m$ values above 99%, indicating nearly perfect alignment with the 24-hour and 168-hour periodic patterns. Furthermore, all datasets maintain $O_m$ values below 3.26%, demonstrating an overall low proportion of outliers.

## 4.2    Validation of LRU Strategy Convergence Optimization

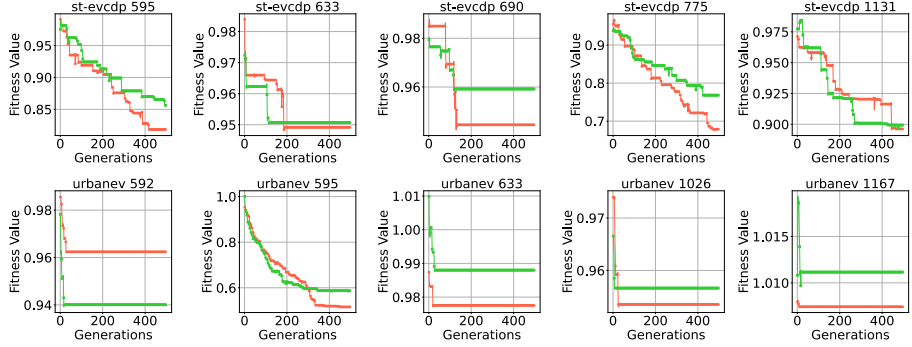

**Fig. 3.** Variation of the best fitness of the population. Due to budget constraints, we only sample 10 regions for evaluation. The experiments used uniform parameters: population size 100, crossover probability 0.8, mutation rate 0.005, and maximum iterations 500. The green curve represents the GA, and the orange curve represents the GA-EVLRU.

To evaluate the impact of LRU on the convergence optimization of GA in battery swapping demand scenarios, we designed systematic ablation experiments with two comparison groups: the basic GA and the GA-EVLRU. Given the extensive geographical coverage of the original datasets (ST-EVCDP and UrbanEV) and the significant variations in demand across regions (some regions exhibit extremely low daily demand that can be met without additional charging), we selected 5 demand-intensive regions from each dataset respectively. Corresponding convergence curves presented in **Fig. 3**.

As shown in **Fig. 3**, except for the urbanev-592, the final convergence value of GA-EVLRU in the other nine test regions consistently lie below the baseline. During the early iterations, the best fitness values of two algorithm fluctuate, but as the process advances, the LRU-based directional mutation guides the search process to favor high-quality gene segments, thereby enabling the GA-EVLRU to approach Pareto optimal solutions more efficiently. To quantitatively evaluate the algorithm performance, three key metrics are selected: Best Fitness $f_{best}$, Generation of Best Fitness $g(f_{best})$, and Mean Best Fitness $\overline{F}_{best}$. Detailed results of 10 test regions presented in the **Table 2**.

**Table 2.** Comparison of GA and GA-EVLRU on two datasets. We use ↓ for the lower the better. The best is boldfaced. The regions and parameters are the same as above.

| Region | GA | | | GA-EVLRU(Ours) | | |
|--------|-----------------|-----------------------|-------------------------------|-----------------|-----------------------|-------------------------------|
| | $f_{best}$ ↓ | $g(f_{best})$ ↓ | $\overline{F}_{best}$ ↓ | $f_{best}$ ↓ | $g(f_{best})$ ↓ | $\overline{F}_{best}$ ↓ |
| *ST-EVCDP* | | | | | | |
| **595** | 0.856 | 491 | 0.907 | **0.818** | 416 | 0.884 |
| **633** | 0.951 | 114 | 0.953 | **0.949** | 184 | 0.955 |
| **690** | 0.959 | 116 | 0.963 | **0.944** | 122 | 0.953 |
| **775** | 0.768 | 422 | 0.837 | **0.674** | 479 | 0.800 |
| **1131** | **0.892** | 476 | 0.923 | 0.896 | 462 | 0.931 |
| *UrbanEV* | | | | | | |
| **592** | **0.940** | 15 | 0.941 | 0.962 | 26 | 0.963 |
| **595** | 0.583 | 410 | 0.656 | **0.516** | 449 | 0.648 |
| **633** | 0.988 | 25 | 0.988 | **0.978** | 18 | 0.978 |
| **1026** | 0.957 | 6 | 0.957 | **0.954** | 23 | 0.954 |
| **1167** | 1.009 | 15 | 1.011 | **1.007** | 7 | 1.007 |

As shown in **Table 2**, it can be seen that in the 8 regions, the GA-EVLRU outperforms the baseline on the $f_{best}$ and the $\overline{F}_{best}$. In 4 regions, the GA-EVLRU leads the baseline on the $g(f_{best})$. Although the baseline shows a faster convergence speed in more regions, it may fall into a local optimal solution due to premature convergence. However, within the limited number of iterations, the final convergence value of the GA-EVLRU lower, so it can more effectively approach the global optimum.

### 4.3    Evaluation of GA-EVLRU Optimization Results

This section employs the GA-EVLRU to optimize the 24-hour charging schedule for swapped batteries, aiming to maximize user satisfaction while minimizing charging costs. Comparative experiments conducted across 10 regions from the ST-EVCDP and UrbanEV datasets evaluate the optimized charging schedules against the immediate swap-and-charge strategy (which charges all swapped-out batteries without delay). The optimal charging schedules for Type A and B batteries are presented in **Fig. 4** respectively. To further quantify the comparison, the detailed results are shown in **Table 3**.

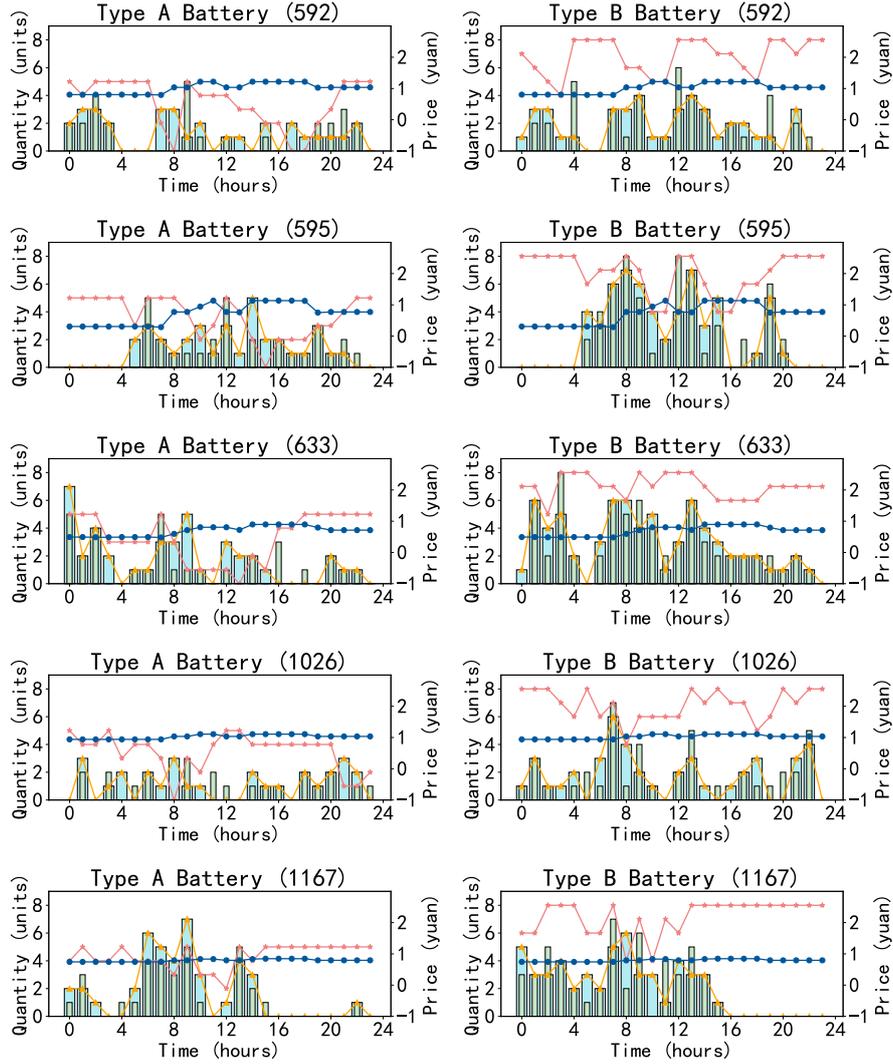

**Fig. 4.** Best A and B Battery Charging Plan on the UrbanEV dataset. The orange line with triangle markers represents the battery swapping demand, the light coral line with asterisk markers represents the fully charged battery inventory; the blue-toned bar chart represents the immediate swap-and-charge strategy, the green-toned bar chart represents the LRU optimization strategy; and the blue line with dot markers represents the electricity price.

**Fig. 4** demonstrates that when the hourly battery demand matches the optimized charging quantity, the fully-charged battery inventory remains stable. The inventory increases when the charging quantity exceeds demand, and decreases in the opposite scenario. Particularly, the optimized plans exhibit distinct price-responsive characteristics, as evidenced in regions 592/595 of UrbanEV, where the algorithm strategically reduces charging activities during peak electricity price periods. Meanwhile, the curve of the number of fully charged batteries is generally higher than user demand, indicating that our plan has always maintained a high level of user satisfaction.

**Table 3.** Results of GA-EVLRU on two datasets. We use ↑ for the higher the better and ↓ for the reverse. $C_{is}$ represents the cost of the immediate swap-and-charge strategy; $C_{ours}$ represents the cost of GA-EVLRU; $r_{opt}$ represents the optimization rate; $\gamma$ represents the user satisfaction level. $\tau$ represents the average iteration time per iteration.

| Region | $C_{is}$ ↓ | $C_{ours}$ ↓ | $r_{opt}$ ↑ | $\gamma$ ↑ | $\tau$/s ↓ |
|--------|-----------|--------------|-------------|------------|------------|
| **ST-EVCDP** | | | | | |
| **682** | 3058.95 | 2988.04 | **2.32%** | 98.57% | 0.556 |
| **690** | 2682.29 | 2524.27 | **5.89%** | 96.67% | 0.569 |
| **775** | 3858.70 | 3713.92 | **3.75%** | 98.21% | 0.555 |
| **895** | 2889.28 | 2550.98 | **11.71%** | 94.64% | 0.554 |
| **1131** | 6469.58 | 5566.46 | **13.96%** | 90.70% | 0.560 |
| **UrbanEV** | | | | | |
| **592** | 5296.59 | 4821.68 | **8.97%** | 94.44% | 0.538 |
| **595** | 5118.65 | 4587.35 | **10.38%** | 95.65% | 0.539 |
| **633** | 4957.69 | 4479.44 | **9.65%** | 90.99% | 0.542 |
| **1026** | 5230.44 | 4724.83 | **9.67%** | 92.96% | 0.540 |
| **1167** | 5098.52 | 4630.88 | **9.17%** | 90.22% | 0.543 |

As shown in **Table 3**, compared with the immediate swap-and-charge strategy, our algorithm achieves cost savings ranging from 2% to 15% across all 10 test regions, with the maximum saving of 13.96% observed in region 1131 of ST-EVCDP. Meanwhile, user satisfaction remains above 90% in all regions, reaching the highest value of 98.57% in region 595 of ST-EVCDP. In addition, our algorithm demonstrates good computational efficiency, with average iteration time kept below 0.6 seconds in all regions. The experimental results convincingly demonstrate the practical applicability and advantages of our algorithm.

## 5    Conclusion

In this paper, nine battery swapping demand datasets are constructed. Empirical evidence shows that the estimated data exhibit good trend stability, basically conform to the periodicity of 24 hours and 168 hours, and the proportion of outliers is all lower than 3.26%, indicating the effectiveness. The GA-EVLRU is proposed, which combines the LRU strategy to guide the search of GA. Compared with the basic GA, it is better in the 8 regions of the constructed ST-EVCDP and UrbanEV datasets. In the final actual solution, compared with the immediate swap-and-charge strategy, the highest optimization cost rate reaches 13.96%. The peak value of user satisfaction reaches 98.57%, and the average single iteration time of the algorithm is within 0.6 seconds.

In the future, the expansibility of improving the GA with the LRU strategy will be considered. We will explore the advantages of GA-EVLRU in solving different problems. At the same time, we will also consider introducing deep learning models that can better capture the changes in battery swapping demands.

**Acknowledgments.** Anonymous submission.

**Disclosure of Interests.** The authors have no competing interests to declare that are relevant to the content of this article.